\documentclass[conference]{IEEEtran}
\IEEEoverridecommandlockouts
\usepackage{cite}
\usepackage{amsmath,amssymb,amsfonts}
\usepackage{algorithmic}
\usepackage{graphicx}
\usepackage{textcomp}
\usepackage{xcolor}

\def\BibTeX{{\rm B\kern-.05em{\sc i\kern-.025em b}\kern-.08em
    T\kern-.1667em\lower.7ex\hbox{E}\kern-.125emX}}
    
\newcommand{\dataset}{$(Net)^2$\hspace{0.01in} }

\makeatletter

\newcommand{\linebreakand}{%
  \end{@IEEEauthorhalign}
  \hfill\mbox{}\par
  \mbox{}\hfill\begin{@IEEEauthorhalign}
}

\def\ps@IEEEtitlepagestyle{
  \def\@oddfoot{\mycopyrightnotice}
  \def\@evenfoot{}
}
\def\mycopyrightnotice{
  {\footnotesize xxx-x-xxxx-xxxx-x/xx/\$31.00~\copyright~2021 IEEE\hfill} 
  \gdef\mycopyrightnotice{}
}

\@ifundefined{showcaptionsetup}{}{
 \PassOptionsToPackage{caption=false}{subfig}}
\usepackage{subfig}
\makeatother

\usepackage{eso-pic}
\newcommand\AtPageUpperMyright[1]{\AtPageUpperLeft{
 \put(\LenToUnit{0.5\paperwidth},\LenToUnit{-1cm}){
     \parbox{0.5\textwidth}{\raggedleft\fontsize{9}{11}\selectfont #1}}
 }}
\newcommand{\conf}[1]{
\AddToShipoutPictureBG*{
\AtPageUpperMyright{#1}
}
}

\makeatother

\author{\IEEEauthorblockN{Onur Barut}
\IEEEauthorblockA{\textit{Dept. of Electrical and Computer Eng.} \\
\textit{University of Massachusetts Lowell}\\
Lowell, USA \\
Onur\_Barut@uml.edu}
\and
\IEEEauthorblockN{Yan Luo}
\IEEEauthorblockA{\textit{Dept. of Electrical and Computer Eng.} \\
\textit{University of Massachusetts Lowell}\\
Lowell, USA \\
Yan\_Luo@uml.edu}
\and
\IEEEauthorblockN{Tong Zhang}
\IEEEauthorblockA{\textit{Network Platforms Group} \\
\textit{Intel Corporation}\\
Santa Clara, USA \\
tong2.zhang@intel.com}
\linebreakand
\IEEEauthorblockN{Weigang Li}
\IEEEauthorblockA{\textit{Network Platforms Group} \\
\textit{Intel Corporation}\\
Shanghai, China \\
weigang.li@intel.com}
\and
\IEEEauthorblockN{Peilong Li}
\IEEEauthorblockA{\textit{Dept. of Computer Science} \\
\textit{Elizabethtown College}\\
Elizabethtown, USA \\
lip@etown.edu}
}

\begin{document}

\title{Multi-Task Hierarchical Learning Based Network Traffic Analytics\\
\thanks{This work was supported in part by Intel Corporation.}
}

\conf{Accepted by International Conference on Communications (ICC21) 14-23 June, 2021} 

\maketitle

\begin{abstract}


Classifying network traffic is the basis for important network applications. Prior research in this area has faced challenges on the availability of representative datasets, and many of the results cannot be readily reproduced. Such a problem is exacerbated by emerging data-driven machine learning based approaches. To address this issue, we present \dataset database with three open datasets containing nearly 1.3M labeled flows in total, with a comprehensive list of flow features, for the research community\footnote{ https://github.com/ACANETS/NetML-Competition2020}. We focus on broad aspects in network traffic analysis, including both malware detection and application classification. As we continue to grow them, we expect the datasets to serve as a common ground for AI driven, reproducible research on network flow analytics. We release the datasets publicly and also introduce a Multi-Task Hierarchical Learning (MTHL) model to perform all tasks in a single model. Our results show that MTHL is capable of accurately performing multiple tasks with hierarchical labeling with a dramatic reduction in training time.
\end{abstract}


\section{Introduction}
Recent advances in technology have provided researchers a lot of data available for analysis. In particular, many studies have been published in various fields including Computer Vision (CV), Natural Language Processing (NLP), and lately Network Traffic Analysis (NTA) investigating big data with Artificial Intelligence approaches \cite{packetcgan_icc20, barut_har, barut_cv, Taylor_2016_eta, Wang_2018_datanet, Yao_2019_attentionLSTM, Gao_2020_cicids2017_nsl_kdd}. Even though many benchmarks and challenges are available for CV \cite{imagenet_cvpr09} and NLP fields \cite{senseval}, there is a lack of common ground for network traffic analysis for researchers to compare their results with others.

Network traffic analysis has become vital for many applications such as quality of service (QoS) control, resource allocation and malware detection. Payload-based packet inspection (DPI) for NTA has serious limitation due to performance, privacy concerns and ineffectiveness on encrypted traffic. Similarly, port-based approach has become obsolete long ago since newer applications mostly use dynamic port allocation. On the other hand, an emerging trend is to employ machine learning (ML) models trained on labeled datasets as they do not rely on port numbers or payload \cite{Gao_2020_cicids2017_nsl_kdd, Ahmad_2018_5ofLuNet, AndersonBlake_2016, CICIDS2017, iscx_vpn2016}. However, each of these methods specialize on a single use-case and cannot be generalized. Therefore, we introduce a multi-task learning scheme to overcome this issue.

There have been a plethora of research attempting to classify and analyze network flows using a variety of datasets. However, unlike the open datasets such as ImageNet \cite{imagenet_cvpr09} available in computer vision research, it is very difficult to find comprehensive datasets for researchers in networking domain to evaluate their proposed techniques, and for the community to reproduce prior art. Many agree that a comprehensive, up-to-date and open dataset for flow analytics is dispensable for the network research community.

\begin{sloppypar}
In this paper, we address a common ground for NTA and introduce \dataset, an open database with a collection of multi-labeled network flows and their features for researchers as a benchmarking platform to evaluate their approaches and contribute to NTA research. In order to be comprehensive, we compile datasets for both malware detection and application type categorization, two representative applications. For malware detection tasks, we introduce a new dataset with raw traffic captures obtained from Stratosphere IPS \cite{stratosphere} website with almost $500k$ network flow samples belonging to $20$ different types of malware and benign classes. We also create another dataset for malware detection which is obtained from the raw traffic captures from well-known CICIDS2017 dataset \cite{CICIDS2017}. We generate around $550k$ flow samples for $7$ different malware types and a benign class. Lastly, we introduce non-vpn2016 dataset with around $163k$ flow samples whose raw traffic data are acquired from ISCX-VPN-nonVPN2016 dataset \cite{iscx_vpn2016} containing different levels of annotations with a number of classes ranging from $7$ to $31$. 

On top of those data, we implement Multi-Task Hierarchical Learning (MTHL) model to perform all the tasks simultaneously as a single classifier. We host the dataset and release the implementations of the algorithms on GitHub and encourage the community to collaborate and grow the collection, pushing the state of art in ML-driven network traffic classification.
\end{sloppypar}

\begin{sloppypar}
Our contribution in this paper is threefold: Firstly, we present a novel malware detection \dataset dataset and curate two other datasets originated from open source packet trace captures (named as CICIDS2017 for malware detection and VPN-nonVPN 2016 dataset for traffic classification) with granulated labels and generate more than $100$ features including flow metadata and protocol-specific features. 
Secondly, we provide an open platform for researchers to address network traffic analytics challenge and evaluate baseline results obtained by RF, SVM and MLP models, facilitating the benchmarking of traffic analytics like ImageNet to CV. Thirdly, we introduce a novel MTHL model for network traffic analysis to perform different tasks with hierarchical labeling.
\end{sloppypar}

\section{Related Work}

\subsection{Malware Detection}
KDD-Cup 99 \cite{KDDcup99} and NSL-KDD datasets \cite{nsl_kdd} are common datasets widely used for malware detection \cite{ensemble_kdd_icc20, ddos_kdd_icc20}. However, those two datasets are out of date and models trained with those are prone to recently emerged attacks. CICIDS2017 dataset \cite{CICIDS2017} is relatively up-to-date and more suitable for detecting recent types of intrusion. Sharafaldin et al. introduce CICIDS2017 dataset with their results obtained using CICFlowMeter \cite{flowmeter} features by comparing several techniques including k-nearest neighbor (kNN), random forest and ID3 and report best F1 score $0.98$ for ID3 algorithm and $0.97$ for random forest whose executing time is one-third of ID3. Gao et al. \cite{Gao_2020_cicids2017_nsl_kdd} compare several machine learning and deep learning techniques including Random Forest, SVM and Deep Neural Networks on NSL-KDD and CICIDS2017 datasets. They perform both binary malware detection and multi-class classification and find out that RF model achieves impressive malware detection accuracy and comparable classification accuracy to other deep learning models in multi-class classification tasks.

\begin{sloppypar}
Random Forest and Support Vector Machine (SVM) classifiers are two effective ways among many machine learning approaches for network traffic analysis \cite{Ahmad_2018_5ofLuNet}. However, most of the researchers use their own data with their own features for NTA, making it difficult to compare one study to another.
\end{sloppypar}

\subsection{Traffic Classification}
Conti et al. \cite{Conti_2016_useractions} use Random forest with their own dataset and their own features and achieve $95\%$ accuracy on user action classification. Taylor et al. \cite{Taylor_2016_eta} conclude that using statistical features of flow vectors of raw packet lengths such as minimum, maximum, mean, median, standard deviation on binary SVM and Random Forest results in $95\%$ accuracy, but when using multi-class classifier, performance drops to $42\%$ for SVM and $87\%$ for RF. Lashkari et al. \cite{iscx_vpn2016} introduce ISCX VPN-nonVPN2016 dataset and analyze it with kNN and C4.5 algorithms.

\begin{sloppypar}
Many researchers adopt VPN-nonVPN2016 dataset and report their results. For example, Yamansavascilar et al. \cite{Baris_2017_iscx} utilize this dataset and compare the results with their own dataset using the same classes. They report that kNN works best for VPN-nonVPN dataset and random forest produces most accurate results on their own dataset. They also report $2\%$ boost in the accuracy of their own dataset with proper feature selection. Wang et al. \cite{Wang_2018_datanet} use packet-level analysis and use deep learning approach for protocol classification. Later, Yao et al. \cite{Yao_2019_attentionLSTM} implement attention-based LSTM (Long Short-Term Memory) to classify protocol types in VPN-nonVPN dataset.
\end{sloppypar}

Previous studies show that the demand for analyzing network flow is increasing and there is not a unique solution for application classification or malware detection in network analysis. Therefore, there is an urgent need of a generic set of features that are known to be useful for malware detection and application classification with a more comprehensive and up-to-date dataset.

\section{\dataset Data Collection}
In this paper, we focus broadly on flow classification tasks for various purposes and at different granularity. We study both malware detection and more general flow type identification. We use "top-level", "mid-level" and "fine-grained" to denote the granularity levels. At "top-level", malware detection is a binary classification problem, i.e. benign or malware, while for flow identification task, the major application categories such as \textit{chat}, \textit{email}, \textit{stream} etc. are annotated. The "mid-level" granularity identifies specific applications such as \textit{facebook}, \textit{skype}, \textit{hangouts} etc. for normal traffic and specific types of malware including \textit{Adload}, \textit{portScan} etc. Finally, the "fine-grained" level performs multi-class classification to identify the detailed types of application traffic such as \textit{facebook\_audio}, \textit{skype\_audio} etc. For malware detection datasets, we do not have fine-grained annotations.

\subsection{Dataset Preparation}
\begin{sloppypar}
We start our work by obtaining publicly available raw packet traces from Stratosphere IPS \cite{stratosphere} and Canadian Institute of Cybersecurity (CIC) \cite{iscx_vpn2016, CICIDS2017}. CIC provides several different types of dataset for Intrusion Detection Systems (IDS) along with an application type classification dataset. Stratosphere IPS only focuses on numerous benign and malware captures. As most of the traffic analysis research resort to extract features from flows or raw packets, we aim to process these packet traces to present a comprehensive list of features including Metadata such as the number of packets and bytes, source and destination port numbers, time duration, time-interval between packets of the flow and protocol-specific features including TLS, DNS and HTTP. Table \ref{data_numbers} give the trace files used to obtain the datasets and the number of flows for each label. We utilize our own accelerated tool to compute these flow features and anonymize the data when making them available.
\end{sloppypar}

\begin{table}[th]
\vspace{-1mm}
\footnotesize
\caption{Number of flow per class for each training set}
\label{data_numbers}
\vspace{-4mm}
\begin{center}
\begin{tabular}{|c|c|c|c|}
\hline
\multicolumn{4}{|c|}{\textbf{\dataset}} \\
\hline
\textbf{fine} & \textbf{mid} & \textbf{top} & \textbf{\# flows} \\
\hline
- & Adload & Malware & 15289 \\
- & Artemis & Malware & 8238 \\
- & Benign & Benign & 76029 \\
- & BitCoinMiner & Malware & 4083 \\
- & CCleaner & Malware & 125637 \\
- & Cobalt & Malware & 1379 \\
- & Downware & Malware & 3857 \\
- & Dridex & Malware & 9208 \\
- & Emotet & Malware & 45907 \\
- & HTBot & Malware & 30442 \\
- & MagicHound & Malware & 31458 \\
- & MinerTrojan & Malware & 4741 \\
- & PUA & Malware & 6171 \\
- & Ramnit & Malware & 400 \\
- & Tinba & Malware & 18627 \\
- & TrickBot & Malware & 37276 \\
- & Trickster & Malware & 57800 \\
- & TrojanDownloader & Malware & 4029 \\
- & Ursnif & Malware & 8482 \\
- & WebCompanion & Malware & 15776 \\
\hline
\multicolumn{4}{|c|}{\textbf{CICIDS2017}} \\
\hline
- & DDoS & Malware & 36136 \\
- & DoS & Malware & 23806 \\
- & benign & benign & 199236 \\
- & ftp-patator & Malware & 3168 \\
- & infiltration & Malware & 53532 \\
- & portScan & Malware & 122430 \\
- & ssh-patator & Malware & 1972 \\
- & webAttack & Malware & 1617 \\
\hline
\multicolumn{4}{|c|}{\textbf{VPN-nonVPN 2016}} \\
\hline
aim\_chat & aim & Chat & 363 \\
bittorrent & bittorrent & P2P & 1215 \\
email & email & Email & 4399 \\
facebook\_audio & facebook & Voip & 35851 \\
facebook\_chat & facebook & Chat & 1340 \\
facebook\_video & facebook & Stream & 336 \\
ftps & ftps & File\_Transfer & 766 \\
gmail\_chat & gmail & Chat & 363 \\
hangouts\_audio & hangouts & Voip & 42913 \\
hangouts\_chat & hangouts & Chat & 2555 \\
hangouts\_video & hangouts & Stream & 1213 \\
icq\_chat & icq & Chat & 377 \\
netflix & netflix & Stream & 413 \\
scp & scp & File\_Transfer & 150 \\
sftp & sftp & File\_Transfer & 175 \\
skype\_audio & skype & Voip & 17668 \\
skype\_chat & skype & Chat & 4526 \\
skype\_files & skype & File\_Transfer & 25533 \\
skype\_video & skype & Stream & 459 \\
spotify & spotify & Stream & 293 \\
tor & tor & Tor\_Browsing & 122 \\
vimeo & vimeo & Stream & 459 \\
voipbuster & voipbuster & Voip & 3821 \\
youtube & youtube & Stream & 731 \\
\hline
\end{tabular}
\end{center}
\vspace{-6mm}
\end{table}

\begin{figure}[tp]
\small
\centering
\includegraphics[width=3in,keepaspectratio]{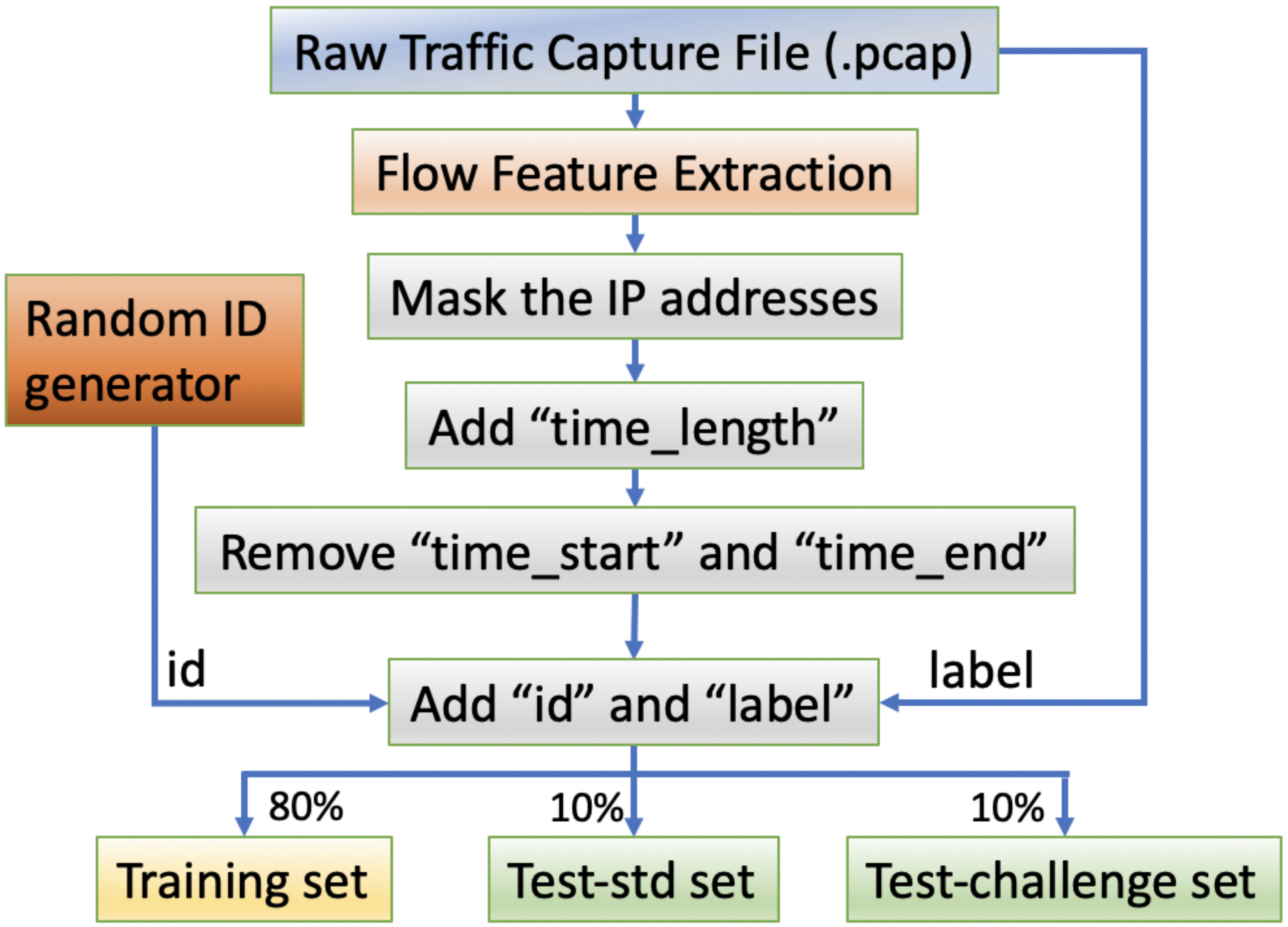}
\caption{Steps of dataset preparation}
\label{data_preparation}
\vspace{-3mm}
\end{figure}

We process the raw packet traces by following the steps depicted in Figure \ref{data_preparation} to generate dataset splits for \dataset database. We design a feature extraction library to compute the flow features by taking up to first $48$ packets in each direction from the raw traffic capture file. Extracted features are stored in JSON format, and features for each flow sample is listed line by line in the output file. Metadata features such as number of packets, source port, destination port etc. can be extracted for every flow sample. However, protocol-specific features such as TLS, DNS and HTTP can be extracted only if the flow sample contains packets for the given protocols. Therefore, some of the flows contain protocol-specific features while all of the flows contain Metadata features. After feature extraction, we mask the source and destination IP addresses. Later, we add \textit{time\_length} as a new feature which is obtained by subtracting \textit{time\_start} from \textit{time\_end}. Then, we remove \textit{time\_start} and \textit{time\_end} from the feature set. These steps are necessary because of two reasons: (1) exposed IP address will reveal anonymity and (2) IP address as well as time features will cause a bias in the prediction due to the data collection setup. After that, "id" and "label" for each flow are added to obtain the \dataset, CICIDS2017 and VPN-nonVPN 2016 datasets. Finally, generated datasets are split into training, test-std and test-challenge subsets using 8:1:1 ratio, respectively. Labels for training set are open to public; however, labels for test-std and test-challenge sets are reserved to evaluate the models of the participants.

\subsection{Malware Detection Datasets}

\subsubsection{\dataset dataset}
This dataset is created for malware detection task by obtaining raw traffic data from Stratosphere IPS. To obtain a dataset which has a similar number of flows with previously released CICIDS2017 \cite{CICIDS2017}, we utilize $19$ different malware $7$ different benign trace files. Different files contain different malware types; therefore, each file is annotated  with its malware type for mid-level annotations. In addition to that, each malware type is also annotated as \textit{Malware} for top-level annotations. 

\subsubsection{CICIDS2017}
This is a set of raw traffic captures for different types of malware attacks and normal flows \cite{CICIDS2017}. Each trace file contains the whole network traffic throughout the day and $5$ traces for each day of a week is collected by CIC. However, only some specific time intervals with packet exchange between the predetermined IP addresses contain the flows we want to extract. Therefore, we download CICIDS2017 dataset from the original source and filter the flows-of-interest according to the descriptions in the dataset website, and extract flow features using the flow feature extraction tool. Similar to the \dataset dataset, top-level annotations in CICIDS2017 contain benign or malware classes while mid-level annotations contain $7$ different malware types and a benign class. 

\subsection{Traffic Classification Dataset} 
The third dataset focuses on application classification and is called VPN-nonVPN 2016. Similar to CICIDS2017, raw traffic capture files are obtained from the CIC website. As each file is named with action performed with the application such as \textit{skype\_audio}, \textit{facebook\_video} etc. we can label those flows in fine-grained annotations to identify $24$ different lower-level classes in many applications. Mid-level annotations contain $18$ type of applications (\textit{facebook}, \textit{skype} etc.) while top-level annotations are $7$ general grouping of those traffic capture data.

VPN-nonVPN dataset contains fine-grained labels whereas \dataset and CICIDS2017 datasets do not reveal lower level actions. Extracted flow numbers from the three datasets are given in Table \ref{data_numbers}.

\section{Methodology}
\label{methodology}

\begin{sloppypar}
In this section we briefly explain the flow features for the training sets of all three datasets. From now on, we only use Training set to conduct our analysis and experiments as only this split will be publicly available with the annotations. We extract four set of features: (1) metadata, (2) TLS, (3) DNS, (4) HTTP. Metadata features are protocol-independent features such as number of packets, bytes inbound and bytes outbound, time length of a flow etc. On the other side, there are protocol-specific features such as TLS, DNS and HTTP. Number of ciphersuites and extensions supported by the client or server are a subset of the TLS features. Similarly, DNS query name and DNS answer IP can be given as example for DNS features. Finally, HTTP code and HTTP method are two examples for HTTP features. While metadata features can be extracted for any kind of flow, protocol-specific features can only be extracted if the given flow contains packets with one of these protocols. Therefore, Table \ref{data_percentage} summarizes number of flows in each training set containing metadata, TLS, DNS and HTTP features.
\end{sloppypar}

\begin{table}[th]
\vspace{-3mm}
\small
\caption{Number of flow features for each training set}
\label{data_percentage}
\vspace{-4mm}
\begin{center}
\begin{tabular}{|l|cccc|}
\hline
& \textbf{Metadata} & \textbf{TLS} & \textbf{DNS} & \textbf{HTTP} \\
\hline
\textbf{\dataset} & 580858 & 135739 & 114522 & 52185 \\
\textbf{CICIDS 2017} & 441897 & 75163 & 93568 & 40314 \\
\textbf{VPN-nonVPN 2016} & 146041 & 1982 & 19362 & 793 \\
\hline
\end{tabular}
\end{center}
\vspace{-3mm}
\end{table}

The imbalance of number of samples in a dataset is an important but mostly inevitable. As an example, in Table \ref{data_numbers}, we observe that \textit{facebook\_audio}, \textit{hangouts\_audio}, \textit{skype\_audio} and \textit{skype\_file} classes are oversampled in the VPN-nonVPN 2016 dataset. Similarly, \textit{CCleaner} in \dataset and \textit{portScan} in CICIDS2017 datasets reveal the same behavior. Such significantly imbalanced class distribution in the training set may introduce a bias to the predictions of the trained models. However, we leave the datasets as they are to offer a challenge for the researchers to tackle with the imbalanced data. 

\subsection{Metadata Features}

Metadata features are mostly composed of statistical features or histograms computed from a flow. Metadata features can be extracted for any flow samples. In this study, $31$ different metadata features are extracted using the flow feature extraction tool. For the histogram-like features such as compact histogram of payload lengths \textit{pld\_ccnt[]} and compact histogram of header lengths \textit{hdr\_ccnt[]}, the returned values are a constant size of array. Unlike histogram arrays, the other metadata features such as number of bytes inbound \textit{bytes\_in} or source port number \textit{src\_prt} return a single value. 

\subsection{Protocol-Specific Features}
Apart from metadata features, TLS, DNS and HTTP protocols are three other sources of features extracted by the feature extraction tool. Several client and server-based TLS-specific features are extracted in our study. Some of these TLS features are the list of offered ciphersuites, the list of advertised extensions and the key exchange length advertised by or and server. In total, $14$ TLS features are extracted. 

Our feature extraction tool returns several DNS query and answer features. Similar to TLS features, there are both single-valued and array-like features in DNS involved flows. For example, \textit{dns\_query\_cnt} and \textit{dns\_answer\_cnt} features are returned as a single integer while the other query and answer features are returned as array of integers for \textit{dns\_answer\_ttl} and strings for the rest. 

For each flow containing HTTP packets, we extract six different HTTP features. \textit{http\_method}, \textit{http\_code} and \textit{http\_content\_len} are the only features which are represented as a single integer value while the others contain string datatype which needs to be processed before classification.

\subsection{Design of Multi-Task Hierarchical Learning}
\begin{figure}[tp]
\centering
\includegraphics[width=\linewidth,keepaspectratio]{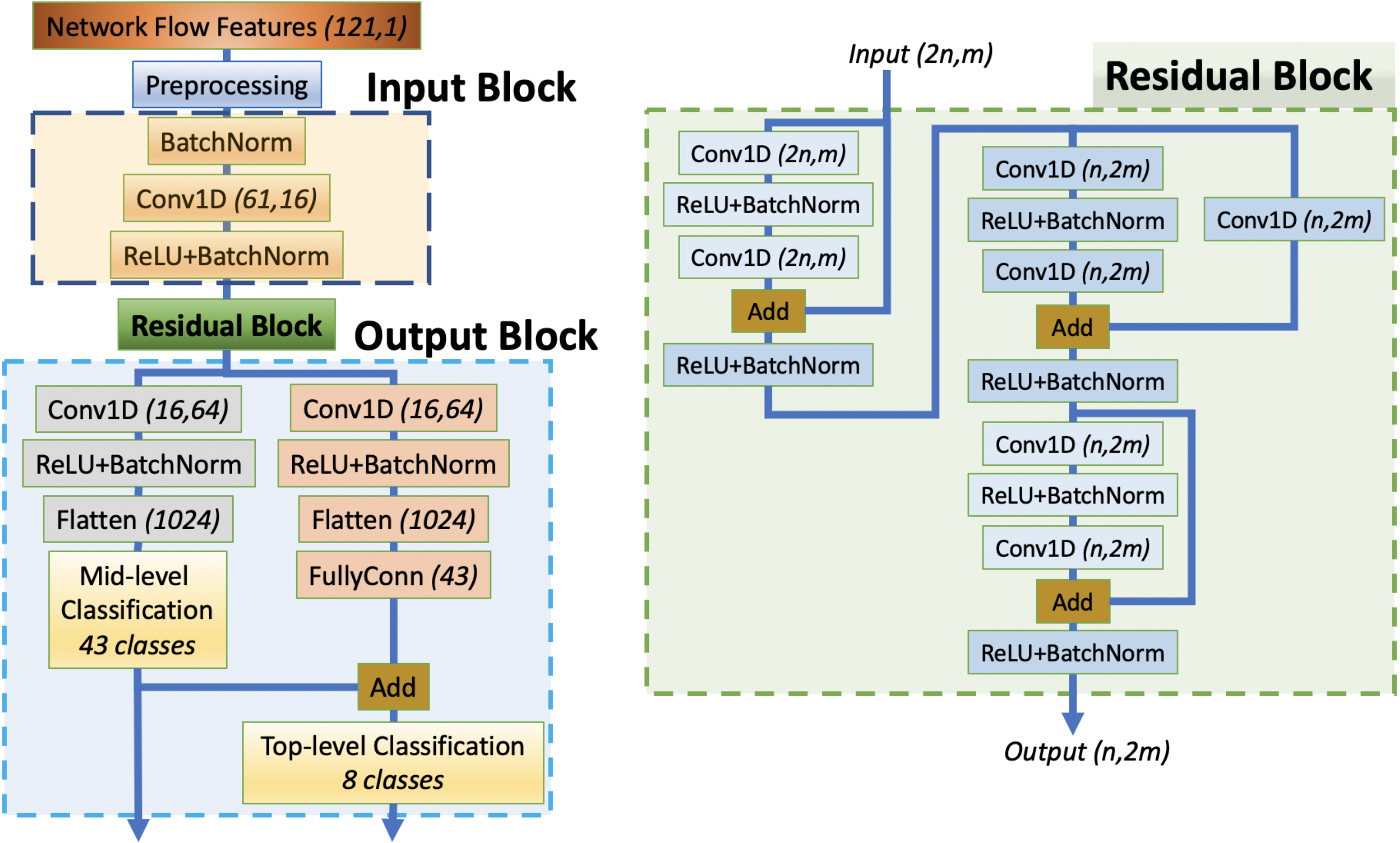}
\caption{Multi-Task Hierarchical Learning (MTHL) model. Residual block is detailed on the right.}
\label{fig:mthl_model}
\vspace{-4mm}
\end{figure}

The structure of MTHL model is provided in Figure \ref{fig:mthl_model}. We now discuss the blocks in detail.

\subsubsection{Preprocessing}
After we obtain the training set split as depicted in Figure \ref{data_preparation}, we need to convert the JSON format data into a two-dimensional array whose columns represent features and rows stand for flow samples. Firstly, we discard source and destination IP address features as they are already masked. All other metadata features in the datasets contain numeric values and hence it is straight-forward to import these datasets into a matrix form that will be used to train the classifiers. For array-like features such as \textit{hdr\_ccnt[]}, we just treat each of those dimensions in this array as a separate feature such as \textit{hdr\_ccnt\_0}, \textit{hdr\_ccnt\_1} etc. up to \textit{hdr\_ccnt\_k} for k+1 dimensional \textit{hdr\_ccnt[]} feature and place each of those into separate columns in the final data matrix. Following this approach yields $121$ columns obtained from $31$ flow metadata features. Finally, the data are normalized so that data in each feature column follows standard normal distribution.

\subsubsection{Input Block}
Batch normalization help the deep model train in a more stable and faster way by normalizing the input batch for the next layer. Therefore, we use batch normalization before any operation in the network. One dimensional convolutional layer is implemented to exploit the spatial correlation between the elements of histogram-like features after batch normalization because these elements are highly correlated. After convolution, Rectified Linear Unit (ReLU) activation function is adopted as it simplifies training the model. Finally, another batch normalization is applied to keep the training stable. Stride of convolution kernel is set to $2$ with zero-padding and $16$ activation maps are obtained in the input block, yielding $61$x$16$ tensor for a single input flow data.

\subsubsection{Residual Block}
Residual connections enables training deeper models and improve the performance. Therefore, instead of a plain neural network, the residual block is implemented whose details are depicted in Figure \ref{fig:mthl_model}. Residual block includes one dimensional convolution, ReLU activation followed by batch normalization, and a residual connection. In the first part, the data shape is preserved as ($61$x$16$) in the convolution operations with stride as $1$. In the second part, spatial size of the data is halved with stride as $2$ and the number of activation maps are doubled from $16$ to $32$. For that purpose, 1D convolution layer is also applied in the residual connection. The last part of the residual block is very similar to the first part. It takes ($32$x$32$) tensor as input and outputs the same shape.

\subsubsection{Output Block}
This block is where the model has multiple branches for different levels of classifications. Both of the branches apply 1D convolution with stride $2$ to further reduce the spatial size and increase the number of activation maps, yielding $16$x$64$ tensor for each flow. After ReLU activation and batch normalization, two-dimensional data is reshaped to one dimension for final classification. Application and malware type classification task, which uses the mid-level annotations, is performed using fully connected layer with softmax activation. On the other hand, top-level classification branch, additional fully connected layer is adopted to enable the connection from the mid-level classification, thus ensuring the hierarchy.

\section{Experimental Evaluation}
In this section, we demonstrate how the proposed granularity of labels in these datasets can be used for Multi-Task Hierarchical Learning (MTHL) to combine all different tasks in a single model and compare its performance with the baselines for flow analysis. 

\subsection{Experiment Setup}
Application and malware type classification tasks are defined as mid-level classification while protocol identification and malware detection are performed as top-level classification tasks. All the flow data from VPN-nonVPN dataset is used with both top-level and mid-level annotations. Since these flows belong to normal traffic, only the malware flows are adopted from \dataset and CICIDS2017 datasets with both mid-level and top-level annotations for the MTHL model.

To evaluate the proposed MTHL model, we prepare four baseline machine learning models for comparison. We utilize Scikit-learn library for traditional machine learning classifier training on Intel(R) Core(TM) i7-6700HQ CPU @ 2.60GHz and Keras library with Tensorflow backend for MTHL model training on NVIDIA GeForce GTX 1060 GPU with 6 GB memory and CUDA version 10.2. 

$80\%$ of the training split is used to train the classifiers while the remaining $20\%$ is reserved to for evaluation and referred as validation set. Macro-average F1 score is utilized to evaluate the performance since it provides more general insight for multi-class classification problems with imbalanced data. Adam optimization with learning rate $0.001$ and cross-entropy are adopted for loss evaluation and training. Batch size is set to $200$ and $100$ the model is trained for $100$ epochs.

\subsection{Baselines}

Random Forest, k-Nearest Neighbor (kNN), Support Vector Machine (SVM) and Multi-layer Perceptron (MLP) are implemented as baselines since they are easy to train and widely used in many applications. As explained in Section \ref{methodology}, protocol-specific features are not available for all of the flows in the datasets. Therefore, we choose only metadata features to expedite the efficiency of the proposed classifiers.

\subsection{Results}
We provide the classification results of $6$ different scenarios obtained using validation set and summarized in Table \ref{baseline_comparison}. Each row gives a different task with the given dataset and the level of the annotation. We observe that different baseline models perform best for each scenario. For example, RF model is the best for application classification dataset while kNN classifier performs the best for \dataset-t and \dataset-m scenarios. Even though CICIDS2017 is another dataset for malware classification, unlike \dataset, the best performing baseline model is MLP for both of the cases. This indicates that there is not a single baseline model that performs best for all the tasks. Moreover, different models would be the best for different tasks, as in our case. However, our proposed MTHL model is capable of executing all the tasks simultaneously and it outperforms the best performing baseline models except for the two tasks. Our model could produce $0.639$ F1 score for malware type classification task in \dataset dataset, which is $0.023$ higher than SVM model but $0.060$ lower than the best performing kNN classifier. Similarly, our model has some limitations while performing application-level classification. RF classifier, which is the best baseline for VPN-nonVPN2016-m task, provides $0.625$ F1 score whereas MTHL could achieve very comparable F1 score with $0.616$.

\begin{table}[th]
\vspace{-2mm}
\footnotesize
\caption{F1 scores for different scenarios. t: top-level, m: mid-level classification task}
\label{baseline_comparison}
\vspace{-4mm}
\begin{center}
\begin{tabular}{|l|c|c|c|c|c|}
\hline
& \textbf{RF} & \textbf{kNN} & \textbf{SVM} & \textbf{MLP} & \textbf{MTHL} \\
\hline
\dataset-t & 0.984 & 0.984 & 0.939 & 0.964 & \textbf{0.994} \\
\hline
\dataset-m & 0.655 & \textbf{0.699} & 0.616 & 0.639 & 0.639 \\
\hline
CICIDS2017-t & 0.978 & 0.989 & 0.989 & 0.989 & \textbf{0.994} \\
\hline
CICIDS2017-m & 0.890 & 0.928 & 0.880 & 0.952 & \textbf{0.955} \\
\hline
VPN-nonVPN2016-t & 0.646 & 0.581 & 0.401 & 0.456 & \textbf{0.661} \\
\hline
VPN-nonVPN2016-m & \textbf{0.625} & 0.576 & 0.423 & 0.388 & 0.616 \\
\hline
\end{tabular}
\end{center}
\vspace{-3mm}
\end{table}

The results in Table \ref{baseline_comparison} also show that the proposed MTHL model can produce more accurate classification results for top-level classes, however, it is slightly less accurate in mid-level class classification. It is important to note that MTHL model is a single model which is trained once to achieve multiple tasks within a single model while all other baseline classifiers are individually trained for the specified task. Moreover, for each task, different baseline models perform the best while our proposed MTHL model can either outperform or operate with comparable performance. For example, RF and kNN models are the best choices for \dataset-t while kNN is the only best classifier for \dataset-m. Similarly, MLP is the best model for CICIDS2017 top and mid-level classification tasks while RF classifier is the most accurate baseline for VPN-nonVPN 2016 top and mid-level classification tasks. Since there is not a single baseline model which performs the best for all of six tasks, one needs to train six tasks with four baselines, making $24$ different training time that can last up to three days of computation. On the other hand, training our MTHL model would take only a few hours on GPU while ensuring a higher performance for four of six tasks. Therefore, we can conclude that MTHL model is capable of learning common representations for different tasks and able to correctly classify the network traffic with different granularity.


\section{Conclusion and Future Work}
\begin{sloppypar}
To enable data driven machine learning based research in network flow analytics, we introduce \dataset datasets curated from open-sources for network traffic classification. We release the flow features and different levels of annotations, aiming to present a common dataset for research community. We also introduce a novel multi-task hierarchical learning model and show that our model can be trained for multiple tasks with different label granularity and achieve higher performance for four of six tasks with much shortened training time when compared to baseline machine learning classifiers.
Our results have inspired potential future work including, utilizing TLS, DNS and HTTP flow features as well as raw bytes of the traffic data to achieve higher accuracy. We expect that the \dataset datasets will promote machine learning based network traffic analytics and establish a common ground for researchers to propose and benchmark their approaches in this AI era.
\end{sloppypar}

\vspace{-1mm}

\section*{Acknowledgments}
\vspace{-1mm}
This work was supported in part by Intel Corporation.

\bibliographystyle{./bibliography/IEEEtran}

\end{document}